\pdfoutput=1

\documentclass[11pt]{article}

\usepackage[]{naacl2021}

\usepackage{times}
\usepackage{latexsym}

\usepackage[T1]{fontenc}

\usepackage[utf8]{inputenc}

\usepackage{microtype}
\usepackage{url}
\usepackage{multirow}
\usepackage{subcaption}
\usepackage{adjustbox}

\title{AdaptSum: Towards Low-Resource Domain Adaptation for \\ Abstractive Summarization}

\author{Tiezheng Yu\thanks{$^*$ Equal contributions. Listing order is random.} , Zihan Liu$^*$, Pascale Fung \\
Center for Artificial Intelligence Research (CAiRE)\\
Department of Electronic and Computer Engineering\\
The Hong Kong University of Science and Technology, Clear Water Bay, Hong Kong\\
\texttt{\{tyuah,zliucr\}@connect.ust.hk},  \texttt{pascale@ece.ust.hk}}

\begin{document}
\maketitle
\begin{abstract}
State-of-the-art abstractive summarization models generally rely on extensive labeled data, which lowers their generalization ability on domains where such data are not available.
In this paper, we present a study of domain adaptation for the abstractive summarization task across six diverse target domains in a low-resource setting.
Specifically, we investigate the second phase of pre-training on large-scale generative models under three different settings: 1) source domain pre-training; 2) domain-adaptive pre-training; and 3) task-adaptive pre-training.
Experiments show that the effectiveness of pre-training is correlated with the similarity between the pre-training data and the target domain task. Moreover, we find that continuing pre-training could lead to the pre-trained model's catastrophic forgetting, and a learning method with less forgetting can alleviate this issue. Furthermore, results illustrate that a huge gap still exists between the low-resource and high-resource settings, which highlights the need for more advanced domain adaptation methods for the abstractive summarization task.\footnote{The code and data are released at: \url{https://github.com/TysonYu/AdaptSum}}
\end{abstract}

\section{Introduction}

Abstractive summarization models aim to extract essential information from long documents and to generate short, concise and readable text. 
Recently, neural abstractive summarization models have achieved remarkable performance~\cite{gehrmann2018bottom,paulus2018a}, and large-scale generative pre-training~\cite{lewis2019bart,raffel2019exploring} has shown itself to be surprisingly effective at generation tasks, including abstractive summarization. However, these models generally require large numbers of human-annotated summaries to achieve state-of-the-art performance, which makes them not scalable to low-resource domains where only a few labeled data are available. 

Domain adaptation methods have naturally arisen to tackle the low-resource issue and enable models to quickly adapt to target domain tasks.
Yet, despite their practicality, very few studies have used domain adaptation methods on the low-resource scenario for the abstractive summarization task.
To address this research gap, we present \textbf{AdaptSum}, the first benchmark to simulate the low-resource domain \textbf{Adapt}ation setting for abstractive \textbf{Sum}marization systems with a combination of existing datasets across six diverse domains (dialog~\cite{gliwa2019samsum}, email~\cite{zhang2019email}, movie review~\cite{wang2016neural}, debate~\cite{wang2016neural}, social media~\cite{kim2019abstractive}, and science~\cite{yasunaga2019scisummnet}), and for each domain, we reduce the number of training samples to a small quantity so as to create a low-resource scenario.

Recently, conducting a second pre-training step on large-scale language models (e.g., BERT~\cite{devlin2019bert}, RoBERTa~\cite{liu2019roberta}) has proven to be effective for domain adaptation tasks~\cite{lee2020biobert,dontstoppretraining2020}. However, the current methods incorporating such a step are mainly focused on classification or classification-based (e.g., named entity recognition) tasks, leaving a research gap in exploring their use for generation tasks. In this paper, we systematically investigate adding a second phase of pre-training on large-scale generative models under three settings: 1) source domain pre-training (SDPT) based on a labeled source domain summarization dataset; 2) domain-adaptive pre-training (DAPT) based on an unlabeled substantial domain-related corpus; and 3) task-adaptive pre-training (TAPT) based on an unlabeled small-scale task-related corpus. 
The second phase of pre-training could cause the catastrophic forgetting in the pre-trained model. Thus, we propose to apply RecAdam~\cite{Chen2020recall} into the pre-training process to alleviate this issue and further improve the adaptation performance.

Experimental results show that SDPT and TAPT can generally improve on the performance of the fine-tuning method, while the effectiveness of DAPT is correlated to the similarity between the pre-training data and the target domain task data. Different from previous insights into adaptive pre-training on classification tasks~\cite{dontstoppretraining2020}, we find that in the summarization task, DAPT could make the adaptation performance worse, even though the pre-training corpus is collected from domain-related sources.
Furthermore, we show that RecAdam can further boost the performance of the second pre-training step by effectively maintaining the pre-trained model's knowledge gained in the first phase of pre-training.

Our contributions are summarized as follows:
\begin{itemize}
    \item We introduce a low-resource domain adaptation scenario for the abstractive summarization task to move towards the fast adaptation of summarization systems.
    \item To the best of our knowledge, we are the first to systematically study the domain- and task-adaptative pre-training for a low-resource generation task.
    \item Our work highlights the research questions and challenges in the low-resource abstractive summarization task, which we hope will catalyze research in this area.
\end{itemize}

\section{Related Work}
\subsection{Abstractive Summarization}
Abstractive summarization aims to generate short, concise and readable text that captures the core meaning of the input documents. Neural networks have achieved remarkable results for the abstractive summarization due to the emergence of Seq2Seq models \cite{sutskever2014sequence} and attention mechanisms \cite{bahdanau2014neural}. \citet{see2017get}, \citet{paulus2017deep} and \citet{gehrmann2018bottom} applied a pointer network to solve the out-of-vocabulary issue.
Further, \citet{see2017get} used a coverage mechanism \cite{tu2016modeling} to keep track of the already summarized content, which discourages repetition, while \citet{paulus2017deep} and \citet{chen2018fast} combined reinforcement learning into an end2end setting. Recently, pre-trained language models \cite{peters2018deep, radford2018improving, devlin2019bert, dong2019unified, lewis2019bart} have achieved impressive gains in a wide variety of natural language tasks. Many studies on the use of pre-trained language models in the abstractive summarization task \cite{liu2019text, yan2020prophetnet, su2020caire, yu2020dimsum} have been undertaken and have achieved the state-of-the-art performance.

\subsection{Domain Adaptation}
Domain adaption for natural language processing and computer vision tasks is widely studied \cite{blitzer2007biographies,mansour2008domain, daume2009frustratingly, sandu2010domain, foster2010discriminative,wang2013domain,sun2016return,liu2019zero,liu2020attention, dontstoppretraining2020,winata2020learning,jadon2020overview,yin2020meta,liu2020zero,liu2020crossner,dai2020modality}.
However, little has been done to investigate domain adaption for the abstractive summarization task.
\citet{hua2017pilot} first studied the adaptation of neural summarization models and showed that the models were able to select salient information from the source domain data. \citet{wang2019exploring} investigated the domain shift problem for the extractive summarization task. Recently, \citet{magooda2020abstractive} studied cross-domain transfer between two entirely different domains and introduced data synthesis methods.
To the best of our knowledge, we are the first to systematically study the domain- and task-adaptative pre-training based on the pre-trained generative model in the low-resource abstractive summarization task across multiple diverse domains.

\begin{table}[t!]
\centering
\begin{adjustbox}{width={0.48\textwidth},totalheight={\textheight},keepaspectratio}
\begin{tabular}{|c|c|c|c|c|c|}
\hline
\multirow{2}{*}{\textbf{Domain}} & \multicolumn{2}{c|}{\textbf{Unlabeled Corpus}} & \multicolumn{3}{c|}{\textbf{Labeled data}} \\ \cline{2-6} 
                        & \# \textbf{Tokens}            & \textbf{Size}           & \textbf{Train}          & \textbf{Valid}          & \textbf{Test}          \\ \hline
Dialog                & 44.96M                 & 212MB           & 300           & 818           & 819          \\ 
Email                   & 117.54M                 & 705MB           & 300           & 1960           & 1906          \\ 
Movie R.            & 11.36M                 & 62MB           & 300           & 500           & 2931          \\ 
Debate                  & 122.99M                 & 693MB           & 300           & 956           & 1003          \\ 
Social M.            & 153.30M                 & 786MB           & 300           & 1000           & 1000          \\ 
Science                 & 41.73M                 & 291MB           & 100           & 350           & 497          \\ \hline
\end{tabular}
\end{adjustbox}
\caption{Data statistics of AdaptSum for the unlabeled corpus and labeled summarization data across the six domains (``R.'' and ``M.'' are the abbreviations for Review and Media, respectively).}
\label{statistics}
\end{table}

\section{AdaptSum}
\label{adaptsum}
\begin{figure}[t!]
    \centering
    \includegraphics[scale=0.51]{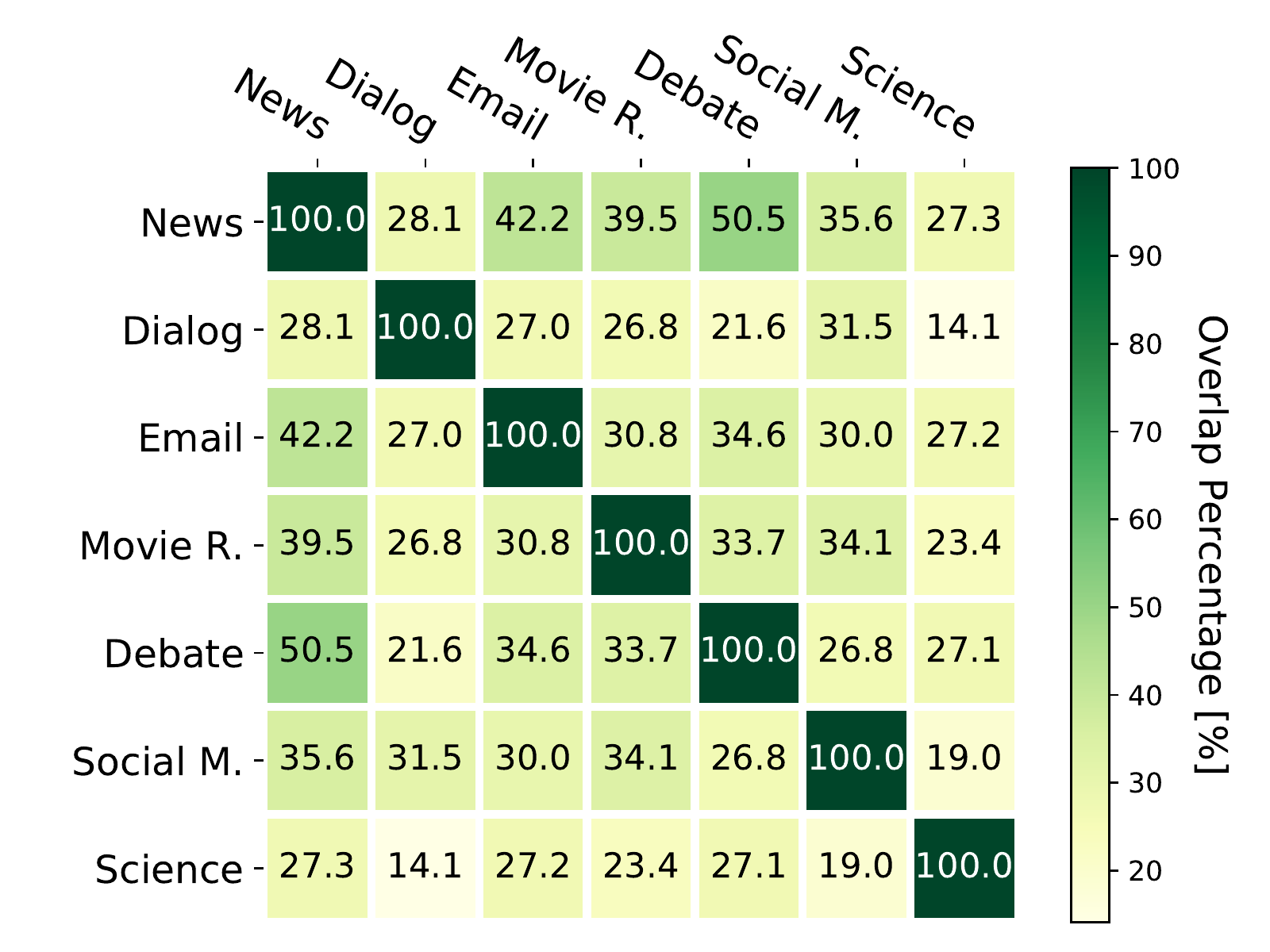}
    \caption{Vocabulary overlaps of the summarization validation set between domains. The News domain is the source domain and the other six domains are low-resource target domains. Vocabularies for each domain are created by considering the top 10K most frequent words (excluding stopwords). We observe that the vocabulary overlaps between domains are generally small, which illustrates that the overlaps between domains are comparably small and the chosen domains are diverse.}
    \label{fig:overlap}
\end{figure}

The goal of AdaptSum is to provide an accessible benchmark for the evaluation of low-resource domain adaptation for abstractive summarization on a diverse set of domains. The vocabulary overlaps between domains are shown in Figure~\ref{fig:overlap}.
AdaptSum consists of six diverse target domains and the corresponding unlabeled domain-related corpora for DAPT. We provide the data statistics of all domains in Table~\ref{statistics}, and the details are as follows.

\paragraph{Dialog}
\citet{gliwa2019samsum} introduced a human-annotated abstractive chat dialog summarization dataset.
The unlabeled dialog corpus from different sources, namely, Reddit conversations,\footnote{\url{https://github.com/PolyAI-LDN/conversational-datasets/tree/master/reddit}} personalized dialogs~\cite{zhang2018personalizing}, empathetic dialogs~\cite{rashkin2019towards}, and Wizard of Wikipedia dialogs~\cite{dinan2018wizard}.

\paragraph{Email}
\citet{zhang2019email} introduced an abstractive business and personal email summarization dataset which consists of email and subject pairs. We collect the unlabeled email corpus from the Enron Email Dataset.\footnote{\url{https://www.cs.cmu.edu/~./enron/}}

\paragraph{Movie Review}
\citet{wang2016neural} introduced a human-annotated abstractive movie review summarization dataset. We collect the unlabeled corpus for this domain from IDMB Movie Review~\cite{maas2011learning}.

\paragraph{Debate} 
\citet{wang2016neural} introduced an abstractive debate summarization dataset which consists of arguments and the debate topic pairs. The unlabeled corpus is from~\citet{ajjour2019data}.

\paragraph{Social Media}
\citet{kim2019abstractive} introduced an abstractive summarization dataset of Reddit TIFU posts, where the summary for each post come from its title. We collect the unlabeled corpus directly from Reddit TIFU.\footnote{\url{https://convokit.cornell.edu/documentation/subreddit.html}}

\paragraph{Science}
\citet{yasunaga2019scisummnet} introduced a human-annotated abstractive summarization dataset on computational linguistics. We collect the unlabeled domain corpus from the ACL anthology~\cite{bird2008acl}.

\section{Methodology}
In this section, we will first introduce the three different settings that we investigate for a second pre-training step. Then, we will discuss how we cope with the catastrophic forgetting issue in the second phase of pre-training.

\subsection{A Second Phase of Pre-Training}
We conduct a second pre-training phase based on a pre-trained generative model, BART~\cite{lewis2019bart}, on three different settings. Then, we fine-tune it to the summarization task in the target domains. The three settings are described as follows.

\paragraph{Source Domain Pre-Training (SDPT)}
Inspired by the cross-domain setting~\cite{jia2019cross,liu2020coach,liu2020crossner}, we leverage substantial training samples from a source (News) domain (XSum~\cite{narayan2018don}), to aid in the fast adaptation in target domains. We choose the News domain as the source domain because it is a rich-resource domain in the summarization task, and from Figure~\ref{fig:overlap}, the similarity between this domain and target domains is generally low which increases the challenge of the domain adaptation.

Our method to conduct SDPT is straightforward. We continue pre-training BART using the source domain summarization data. The objective function for this pre-training is not the sentence reconstruction, as in the original pre-training of BART. Instead, we utilize the supervisions from the source domain summarization data to train BART on the summarization task.
The purpose of this pre-training is to inject the task knowledge into the pre-trained language model so that the model can quickly adapt to the same task in target domains.

\paragraph{Domain-Adaptive Pre-Training (DAPT)}
We leverage an unlabeled domain-related corpus to continue pre-training BART using its original pre-training objective function (corrupting documents and then optimizing a reconstruction loss—the cross-entropy between the decoder’s output and the original document).
The intuition behind this method is to introduce the domain knowledge into the pre-trained language model so as to enable its fast adaptation to the target domains.

\paragraph{Task-Adaptive Pre-Training (TAPT)}
The size of the domain-related corpus for DAPT is usually enormous, which results in two potential drawbacks. First, such a large corpus might not be always available, especially for the low-resource domains. Second, pre-training on such a large corpus is time-consuming and requires excessive computational resources.
Therefore, investigating pre-training on a smaller unlabeled corpus is a practical and beneficial research direction.
TAPT refers to pre-training on a set of the unlabeled documents in the target domain's summarization task. Compared to DAPT, TAPT uses a much smaller but far more task-relevant pre-training corpus since it directly uses the input documents from summarization task. This setting makes TAPT much less expensive to run and independent of the collection of the large domain-related corpus.

\begin{table*}[]
\centering
\begin{adjustbox}{width={0.96\textwidth},totalheight={\textheight},keepaspectratio}
\begin{tabular}{l|cccccc|c}
\hline
\multicolumn{1}{c|}{\textbf{Models}} & \textbf{Dialog} & \textbf{Email} & \textbf{Movie R.} & \textbf{Debate} & \textbf{Social M.} & \textbf{Science} & \textbf{Average} \\ \hline \hline
BART Fine-tuning                    & 39.95  & 24.71 & 25.13    & 24.48  & 21.76     & 72.76   & 34.80   \\ \hline
SDPT                         & 42.84  & 25.16 & 25.45    & 25.61  & 22.43     & \textbf{73.09}   & 35.76   \\
\quad w/ RecAdam                   & \textbf{45.23}      & \textbf{26.97}     & \textbf{26.06}        & 25.17      & \textbf{23.25}         & 72.60       & \textbf{36.55}       \\
DAPT                         & 41.22  & 26.50 & 24.25    & \textbf{26.71}  & 22.95     & 71.88   & 35.59   \\
\quad w/ RecAdam                   & 40.05      & 25.66 & 25.78        & 25.01      & 21.51         & 72.23       & 35.04       \\
TAPT                         & 40.15  & 25.30 & 25.27    & 24.59  & 22.81     & 73.08   & 35.20   \\
\quad w/ RecAdam                   & 41.34  & 25.73 & 25.65    & 24.70  & 23.01     & 72.80   & 35.54   \\ \hline
\end{tabular}
\end{adjustbox}
\caption{ROUGE-1 scores on different pre-training methods compared to the baseline BART over all domains.}
\label{tab:results}
\end{table*}

\begin{table*}[]
    \centering
    \begin{adjustbox}{width={0.87\textwidth},totalheight={\textheight},keepaspectratio}
    \begin{tabular}{c|cccccc|c}
    \hline
    \multicolumn{1}{c|}{\textbf{Corpus}} & \textbf{Dialog} & \textbf{Email} & \textbf{Movie R.} & \textbf{Debate} & \textbf{Social M.} & \textbf{Science} & \textbf{Average} \\ \hline \hline
    DAPT        & 212MB  & 705MB & 62MB   & 693MB  & 786MB     & 291MB   & 458.2MB   \\
    TAPT        & 7.9MB  & 14MB & 3.3MB    & 2.4MB  & 74MB     & 384KB   & 17.0MB   \\ \hline
    \end{tabular}
    \end{adjustbox}
    \caption{Corpus size comparisons between DAPT and TAPT.}
    \label{tab:corpus_size}
\end{table*}

\subsection{Recall and Learn}
Although the second pre-training step allows the pre-trained model to learn the task or domain knowledge, it might lead to the catastrophic forgetting issue and cause the pre-trained model to partly lose the language understanding ability that it gains in the first pre-training step. To alleviate this issue, we expect the pre-trained model to recall the previously learned knowledge during the process of learning new knowledge. A straightforward way to achieve this goal is to borrow the idea of continual learning methods~\cite{kirkpatrick2017overcoming,lopez2017gradient,Chen2020recall}. In this paper, we adopt RecAdam from~\citet{Chen2020recall} for the second phase of pre-training to weaken the catastrophic forgetting issue. The reason for choosing RecAdam is twofold: 1) it does not require the first step pre-training data from the pre-trained model, which is usually not available; 2) it is the most recent approach that is being successfully applied to natural language processing tasks. The RecAdam is introduced as follows.

Based on the Adam optimizer~\cite{AdamKingma2015}, RecAdam reconstructs the objective function to allow it to gradually shift to the target task:
\begin{equation}
    Loss = \lambda(t) \cdot Loss_{T} + (1-\lambda(t)) \cdot Loss_{S}, \label{eq_loss}
\end{equation}
\begin{equation}
    \lambda(t) = \frac{1}{1+ \textnormal{exp}(-k \cdot (t-t_0))}, \label{eq_lambda}
\end{equation}
where $k$ and $t_0$ are the hyper-parameters controlling the annealing rate and time steps, $Loss_{T}$ represents the target task objective function, and $Loss_{S}$ is used to simulate the first pre-training step of the pre-trained model. $Loss_{S}$ can be simplified as:
\begin{equation}
    Loss_{S} = \frac{1}{2} \gamma \sum_{i} (\theta_i - \theta^*_i)^2,
\end{equation}
where $\frac{1}{2} \gamma$ is the coefficient of the quadratic penalty, $\theta$ is the parameters of the model, and $\theta^*$ (fixed) is the original parameters of the pre-trained model.

Although RecAdam has shown its effectiveness in fine-tuning BERT-like models (e.g., BERT~\cite{devlin2019bert} and ALBERT~\cite{Lan2020ALBERT}) to the GLUE benchmark~\cite{wang2018glue}, exploring the effectiveness of RecAdam in the second phase of pre-training for generative pre-trained models is not trivial. First, the second pre-training step of a language model is a completely different task compared to fine-tuning to downstream tasks. Second, a generative model (e.g., BART) is structurally different from BERT-like models. Third, the corpus sizes for SDPT and DAPT are generally much larger than the sizes of GLUE tasks, which could affect the learning process.


\section{Experimental Setup}
\paragraph{Training Details}
We evaluate all of our models on AdaptSum. For the dialog and email domains, we use the standard splits of \cite{gliwa2019samsum, zhang2019email}, while for movie review, debate, social media and science domains, we split the whole dataset into training, validation and test sets by ourselves since the original works do not specify how to split these datasets or the published datasets do not contain the split training, validation and test sets.
Since the dataset sizes are limited for science, movie review and dialog domains, the maximum training samples for these domains are 100, 300, and 300, respectively, while for dialog, email, and social media domains, the maximum training samples for them are 14732, 14436, and 60354, respectively, and we select 300 samples for each domain to construct a low-resource setting.
We truncate the input documents into 1024 tokens due to the limitation of the maximum input length for BART.
For all the experiments, we use the BART-base version to implement our models. We use a mini-batch size of 4 with a gradient accumulation for 10 iterations. We use Adam optimizer with momentum $\beta_1=0.9$, $\beta_2=0.998$ and noam decay with warm up steps of 1000. In the decoding stage, we use beam search with a beam size of 4. The decoding process will not stop until an end-of-sequence (EOS) token is emitted or the length of the generated summary reaches to 256 tokens. As for the hyperparameters of RecAdam, we select the best $t0$ and $k$ in $\{500, 600, 700, 800, 900, 1,000\}$ and $\{1e-2, 1e-3, 1e-4, 1e-5, 1e-6\}$, respectively, for the annealing coefficient $\lambda(t)$ (Eq.~\ref{eq_lambda}).

\paragraph{Baseline} 
As our baseline, we use an off-the-shelf BART model~\cite{lewis2019bart} and perform supervised fine-tuning of its parameters for the summarization task in each domain. BART serves as a good baseline since it provides the state-of-the-art performance in the summarization task. And, as a single generative language model, it can be easily adapted to different target domains.

\paragraph{Evaluation Metrics}
We use ROUGE \cite{lin2003automatic} to measure the quality of the summary produced in our experiments. Following the previous work~\cite{nema2017diversity}, we report ROUGE F1 (ROUGE-1) on the AdaptSum dataset.\footnote{We use {\fontfamily{qcr}\selectfont pyrouge} to compute all ROUGE scores, with parameters ``-c 95 -2 4 -U -r 1000 -n 4 -w 1.2 -a''. The full results of all the models with ROUGE-2 and ROUGE-L are reported in the Appendix.}

\section{Results \& Analysis}
\subsection{Main Results}
From Table~\ref{tab:results}, we can see that SDPT is able to generally improve the summarization performance of the fine-tuning method for all domains. This is because SDPT teaches the model how to do the task using large numbers of annotated examples, which enables the model to adapt to target domains faster than the fine-tuning method, and SDPT is able to outperform both DAPT and TAPT in terms of the averaged ROUGE-1 score.
The enormous unlabeled corpus makes DAPT quite effective in certain domains, such as email, debate and social media, with close to or more than 2 ROUGE-1 scores improvements over the fine-tuning baseline. As we can see from Table~\ref{tab:corpus_size}, although TAPT uses a far smaller pre-training corpus than DAPT, the performance of TAPT is on par with that of DAPT, which accords with the results in~\citet{dontstoppretraining2020}, where the experiments are conducted for domain adaptation in classification tasks. 
Additionally, adding RecAdam into the second phase of pre-training can generally further boost the adaptation performance for SDPT and TAPT, while it only boost the performance on the movie review and science domains for DAPT. 
We conjecture that a relatively large corpus can potentially weaken the effectiveness of RecAdam, and we observe that the corpus used for DAPT is comparably small for movie review and science domains, and the number of data samples for XSum (204k) is also much smaller than those of DAPT corpora in many domains (e.g., email), which have more than 1M sentences.
According to Eq.~\ref{eq_loss}, extensive training data could result in a comparatively large $Loss_s$ (the model's parameters tend to be greatly modified) which lead to an unstable loss and a negative effect to the pre-training process. In addition, we find that RecAdam is originally shown to be effective at fine-tuning to the downstream GLUE tasks~\cite{Chen2020recall}, the sizes of which are much smaller than the datasets used for SDPT and DAPT.

\begin{table}[]
    \centering
    \begin{adjustbox}{width={0.46\textwidth},totalheight={\textheight},keepaspectratio}
    \begin{tabular}{c|cc}
    \hline
    \textbf{Domains}   & \textbf{DAPT Corpus} & \textbf{TAPT Corpus} \\ \hline
    Dialog    & 37.56 (-7.04)       & 44.60       \\
    Email     & 51.87 (-5.93)      & 57.80       \\
    Movie R.  & 46.63 (\textbf{-14.59})      & 61.22       \\
    Debate    & 53.49 (-8.99)      & 62.48       \\
    Social M. & 48.10 (-3.82)      & 51.92       \\
    Science   & 36.94 (\textbf{-20.90})      & 57.84       \\ \hline
    Average   & 45.44 (-10.54)     & 55.98 \\ \hline
    \end{tabular}
    \end{adjustbox}
    \caption{Vocabulary overlaps (\%) between the pre-training corpus (for DAPT or TAPT) and the validation set of the summarization task for each domain. The numbers in the brackets denote the vocabulary overlap differences between the two pre-training corpora, and the bold numbers denote the large discrepancies.}
    \label{tab:corpus_task_overlap}
\end{table}

\subsection{How Pre-training Data Affects DAPT}
According to prior experiments on domain adaptation for classification or classification-based tasks~\cite{beltagy2019scibert,lee2020biobert,dontstoppretraining2020}, DAPT improves the performance for all domains on the fine-tuning baseline. However, as we can see from Table~\ref{tab:results}, DAPT causes the performance to drop for the movie review and science domains in the summarization task, while TAPT boosts the performance for all the domains. To further investigate the reasons, we aim to analyze the similarity (e.g., vocabulary overlap) between the pre-training corpus for DAPT and the summarization task in the target domain, which we represent with the target domain validation set of the summarization task to represent. We notice that 
it is difficult to justify how much overlap is large enough for DAPT to be considered as effective. Hence, we add the TAPT corpus, which is directly related to the target domain's summarization task, as an upper bound for the comparison.

Table~\ref{tab:corpus_task_overlap} illustrates the vocabulary overlaps for DAPT and TAPT for each domain.\footnote{To ensure the comparison between DAPT and TAPT is fair, we sample partial data from the DAPT corpus to make its size comparable to the TAPT corpus and create vocabularies for each based on the top 5K most frequent words (excluding stopwords). The vocabulary for the validation set of the summarization task is also created in the same way.} We find large discrepancies between the DAPT corpus and TAPT corpus on the movie review and science domains, which indicates that the domain-related corpora in these two domains are not quite related to the task domains, and pre-training on a domain-unrelated or less related corpus can lead to a performance drop compared to the fine-tuning method. Given that the corpus construction is done by looking for the domain-related sources (as mentioned in Section~\ref{adaptsum}), the experimental results point out that collecting a domain-related corpus for DAPT in the summarization task is not straightforward. \textit{Thus, we leave exploring how to construct an effective corpus for DAPT for future work.}

\begin{figure}
    \centering
    \includegraphics[scale=0.48]{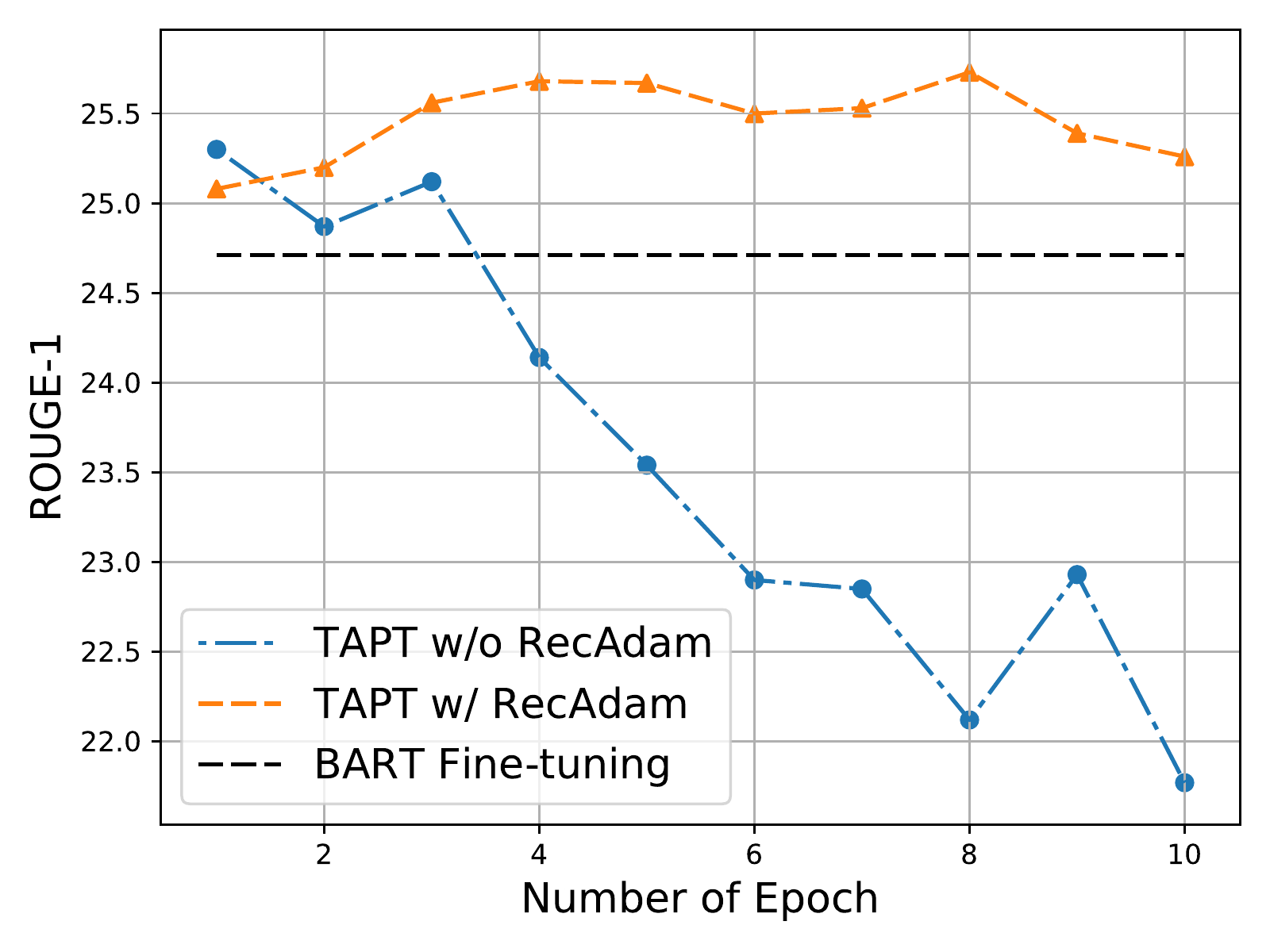}
    \caption{TAPT performance over different pre-training epoch numbers in the email domain in terms of using and not using RecAdam.}
    \label{fig:tapt_through_epoch}
\end{figure}

\begin{table*}[]
\centering
\begin{adjustbox}{width={0.93\textwidth},totalheight={\textheight},keepaspectratio}
\begin{tabular}{c|cc|cccccc}
\hline
\multirow{2}{*}{} & \multicolumn{2}{c|}{\textbf{Source Domains}} & \multicolumn{6}{c}{\textbf{Target Domains}}                        \\ \cline{2-9} 
                  & \textbf{XSum}             & \textbf{CNN/DM}           & \textbf{Dialog} & \textbf{Email}  & \textbf{Movie R.} & \textbf{Debate} & \textbf{Social M.} & \textbf{Science} \\ \hline
Document    & 354.16           & 676.03           & 91.64  & 124.47 & 2112.97 & 196.69 & 229.31    & 633.03  \\
Summary    & 21.13            & 57.91            & 20.28  & 4.10   & 21.28    & 11.07  & 6.31      & 150.01  \\ \hline
\end{tabular}
\end{adjustbox}
\caption{Averaged length of the input documents and output summaries for the source and target domains.}
\label{tab:length}
\end{table*}

\subsection{Catastrophic Forgetting Issue}
We speculate that the second phase of pre-training will result in the catastrophic forgetting for the pre-trained model, which could hurt the adaptation performance. 
Figure~\ref{fig:tapt_through_epoch} illustrates that the performance of TAPT without RecAdam keeps dropping as the pre-training continues, and it starts to perform worse than the fine-tuning method after three epochs' pre-training, while the performance of TAPT with RecAdam remains stable at around a 25.5 ROUGE-1 score. 
We conjecture that excessive pre-training makes the pre-trained model overfit to the pre-training data and partially lose its language understanding and generation ability. However, the model is required to possess both language ability and domain knowledge for better performance in the domain adaptation task.
RecAdam helps the pre-trained model preserve its original language ability while continuing pre-training on a new corpus, which boosts the effectiveness of pre-training.
However, as we can see from Table~\ref{tab:results}, RecAdam fails to improve the performance on DAPT using large corpora. We speculate that the catastrophic forgetting issue does not do much harm to the performance of DAPT because pre-training on the large corpus enables the pre-trained model to possess a good language understanding ability in the target domain even though it could lead to partial forgetting in previous domains, and RecAdam makes DAPT stay somewhere in the middle (not forgetting much the previous learned knowledge, but not learning well in the target domain, either).
\textit{It indicates that more advanced learning methods are needed for coping with the second pre-training phase on a large corpus.}

\begin{table}[]
\centering
\begin{adjustbox}{width={0.49\textwidth},totalheight={\textheight},keepaspectratio}
\begin{tabular}{c|cccc}
\hline
\textbf{Domains}   & \textbf{BART}   & \textbf{SDPT}  & \textbf{DAPT}  & \textbf{SDPT+DAPT} \\ \hline
Dialog    & 39.95            & \textbf{42.84} & 41.22 & 42.27     \\
Email     & 24.71            & 25.16 & \textbf{26.50} & 23.71     \\
Movie R.  & 25.13            & \textbf{25.45} & 24.25 & 22.20     \\
Debate    & 24.48            & 25.61 & \textbf{26.71} &  25.16        \\
Social M. & 21.76            & 22.43 & \textbf{22.95} &  22.03         \\
Science   & 72.76            & \textbf{73.09} & 71.88 & 71.56     \\ \hline
Average   & 34.80            & \textbf{35.76} & 35.59 & 34.49          \\ \hline
\end{tabular}
\end{adjustbox}
\caption{ROUGE-1 results for SDPT+DAPT compared to the SDPT, DAPT and BART fine-tuning.}
\label{tab:sdpt_dapt}
\end{table}

\subsection{Incorporating SDPT and DAPT}
Intuitively, incorporating both the summarization task and target domain knowledge into the pre-trained model could further boost the domain adaptation performance in the summarization task. Therefore, we propose to combine SDPT and DAPT in the second pre-training step. Since SDPT and DAPT use different objective functions, jointly learning these two tasks will make BART confused about what to generate (summarization or sentence reconstruction) given the input sequences. To cope with this issue, we use two BART models (one for SDPT and one for DAPT) and share their encoders in this joint pre-training process to learn the knowledge from both the task and domain. Then, we use the BART model for SDPT to fine-tune to the summarization task in the target domain. 

As shown in Table~\ref{tab:sdpt_dapt}, the experimental results are contradictory to the intuition. We find that SDPT+DAPT can not further improve upon the performance of SDPT and DAPT. For the dialog and social media domains, the performances of SDPT+DAPT stay between those of SDPT and DAPT, while for the science, movie review and email domains, the performances of SDPT+DAPT are even lower than that of the BART fine-tuning. We conjecture that SDPT and DAPT are two completely different tasks, and jointly pre-training based on them could confuse the model about the knowledge that it learns. \textit{However, integrating the task and domain knowledge is still a promising direction for domain adaptation. We leave how to incorporate SDPT and DAPT for future work.}

\begin{table}[]
\centering
\begin{adjustbox}{width={0.48\textwidth},totalheight={\textheight},keepaspectratio}
\begin{tabular}{c|ccc}
\hline
\textbf{Domains} & \textbf{BART} & \textbf{SDPT (XSum)} & \textbf{SDPT (CNN)} \\ \hline
Dialog           & 39.95         & 42.84                & 43.13               \\
Email            & 24.71         & 25.16                & 23.81               \\
Movie R.         & 25.13         & 25.45                & 24.51               \\
Debate           & 24.48         & 25.61                & 23.98                    \\
Social M.        & 21.76         & 22.43                & 22.56                    \\
Science          & 72.76         & 73.09                & 72.41               \\ \hline
Average          & 34.80         & 35.76                & 35.07                    \\ \hline
\end{tabular}
\end{adjustbox}
\caption{ROUGE-1 results for SDPT based on the XSum and CNN/DM (denoted as CNN in the table) datasets.}
\label{tab:sdpt}
\end{table}

\begin{figure*}
\centering
\begin{subfigure}{.49\textwidth}
    \centering
    \includegraphics[scale=0.535]{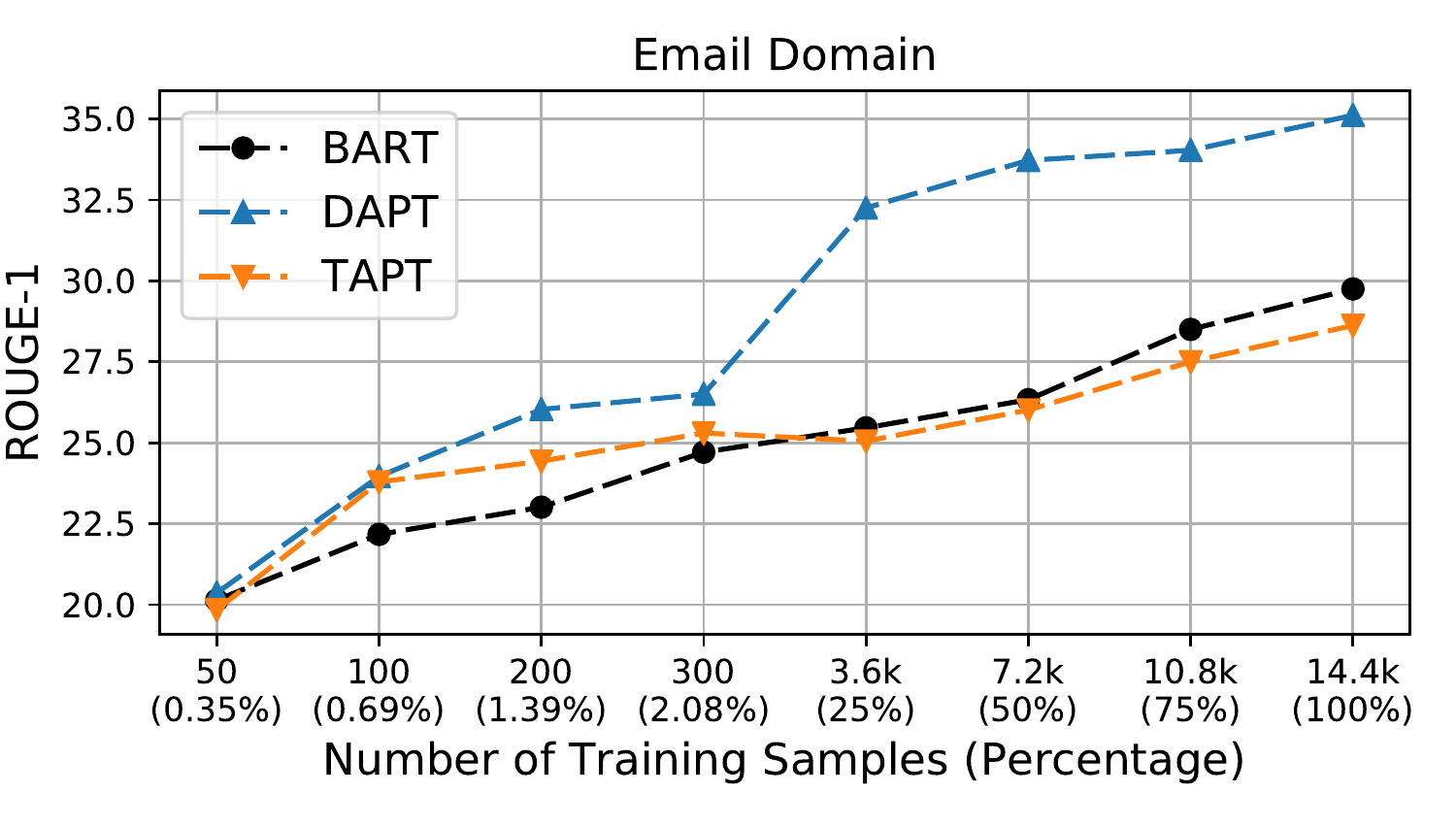}
\end{subfigure}
\begin{subfigure}{.49\textwidth}
    \centering
    \includegraphics[scale=0.535]{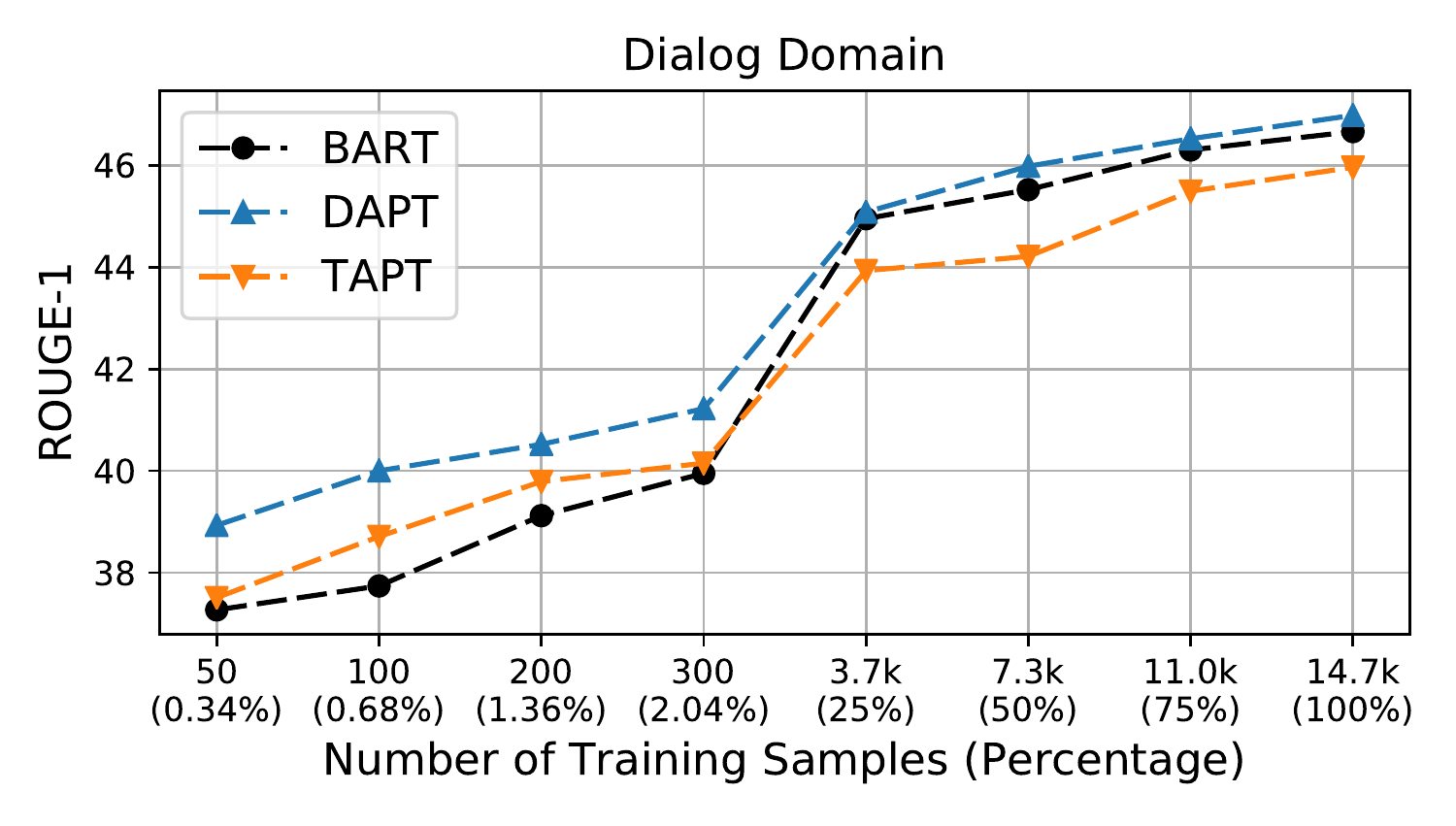}
\end{subfigure}
\caption{ROUGE-1 results of BART fine-tuning, DAPT and SDPT over different numbers of training data for email (left) and dialog (right) domains. We consider both low-resource settings (50, 100, 200 and 300 ($\sim$2\%) samples), medium-resource settings (25\% and 50\% samples), and high-resource settings (75\% and 100\% samples).}
\label{fig:dff_num}
\end{figure*}

\subsection{Different Source Domain Data for SDPT}
To explore how different source domain data can affect the performance of SDPT, we use another News domain dataset, CNN/Daily Mail (DM) dataset~\cite{hermann2015teaching,nallapati2016abstractive}, as the labeled summarization data for SDPT.
As we can see from Table~\ref{tab:sdpt}, SDPT based on CNN/DM only achieves marginal improvements upon the BART fine-tuning baseline in terms of the averaged score, and for all the domains, it generally performs worse or similar compared to SDPT based on XSum. Since both of them are from the News domain but the number of training samples in CNN/DM (287k) is higher than that in XSum (204k), pre-training on CNN/DM should have achieved better performance than pre-training on XSum. To further analyze the reason, we calculate the averaged length of input documents and output summaries for the source and target domains. From Table~\ref{tab:length}, we find that the averaged length of XSum is much shorter than that of CNN/DM in terms of both document and summary, and surprisingly, SDPT based on XSum can outperform SDPT based on CNN/DM in domains with short length document and summary (e.g., debate and email) as well as the domains with long length document or summary (e.g., movie review and science). Hence, we conjecture that pre-training with relatively short document and summary is more effective for SDPT. Another reason can be attributed to the fact that the summaries of the CNN/DM tend to copy the content in the input documents, while XSum has larger amounts of novel tokens in the summaries. Therefore, we conjecture that XSum enables model learn a more powerful summarization ability, which helps it to better adapt to low-resource target domains. \textit{We leave investigating the effectiveness of different source domain datasets in SDPT for future work.}

\subsection{Performance vs. Training Sample Size}
We investigate how well models perform in an extremely low-resource scenario (e.g., 50 training samples) and the performance discrepancies among different levels of resources.
The performance over different numbers of training samples is illustrated in Figure~\ref{fig:dff_num}. 
We find that BART fine-tuning with the 25\% data samples significantly outperforms that with $\sim$2\% data samples in the dialog domain, but such improvements are not remarkable in the email domain. We conjecture that the input and output lengths for the email domain are relatively short compared to the dialog domain (according to Table~\ref{tab:length}), making the domain adaptation easier.

Interestingly, DAPT outperforms other models in the medium-resource and high-resource settings in the email domain but not in the dialog domain. We speculate the reasons are twofold. First, based on the vocabulary overlaps from Table~\ref{tab:corpus_task_overlap}, the email corpus is more effective for DAPT than the dialog domain. Second, email corpus is much larger than the dialog corpus from Table~\ref{tab:corpus_size}. However, the performance of DAPT using a high-quality corpus will be still limited by the low-resource scenario, and it needs large enough training samples to achieve remarkable improvements.
Moreover, the performance of TAPT is better than BART fine-tuning in the low-resource setting, while it becomes worse in the medium-resource and high-resource settings. We conjecture that training with more data will aggravate the catastrophic forgetting caused by TAPT, which leads to the worse performance.

Surprisingly, the performance of DAPT with medium-resource is close to that with high-resource, which can be attributed to the combination of the powerful adaptation ability of the large pre-trained generative model and the effectiveness of the second phase of pre-training.
However, there is still a large performance gap for all the models between the low-resource and high-resource settings and all the models perform badly when there is only 50 training samples, \textit{which highlights the needs for more advanced domain adaptation models for the summarization task.}

\section{Conclusion and Future Work}
In this paper, we present AdaptSum, the first benchmark to simulate the low-resource setting for the abstractive summarization task with a combination of existing datasets across six diverse domains. We systematically study three different methods for a second phase of pre-training (i.e., SDPT, DAPT and TAPT), and propose to leverage RecAdam to alleviate the catastrophic forgetting issue caused by the continuing pre-training.
Experiments show that SDPT and TAPT can generally improve on the performance of the fine-tuning method, while the effectiveness of DAPT depends on the similarity between the pre-training data and the target domain task data, which is different from the insights into DAPT for classification tasks. Further analysis illustrates that RecAdam successfully alleviates the catastrophic forgetting issue for TAPT and further boost its performance. 

Finally, our work highlights several research challenges in low-resource domain adaptation for the abstractive summarization task: (1) How to construct an effective corpus for DAPT; (2) How to better cope with the catastrophic forgetting issue for the second pre-training phase on a large corpus; (3) How to effectively integrate the task and domain knowledge (i.e., incorporate SDPT and DAPT); (4) How to choose better source domain datasets for conducting SDPT; (5) How to build a more powerful domain adaptation models for the extremely low-resource summarization task. We hope that the proposed dataset and the highlighted research directions will accelerate the studies in this area.

\section*{Acknowledgments}
We want to thank the anonymous reviewers for their constructive feedback. This work is partially funded by ITF/319/16FP, ITS/353/19FP and MRP/055/18 of the Innovation Technology Commission, the Hong Kong SAR Government.

\bibliography{anthology,custom}
\bibliographystyle{acl_natbib}

\appendix

\section{Full Results of All Models}
\label{sec:appendix A}
The full results of all models are shown in Table~\ref{tab:all_results}. 

\begin{table*}[h]
\begin{adjustbox}{width={1\textwidth},totalheight={\textheight},keepaspectratio}
\begin{tabular}{c|cccccccc}
\hline
\textbf{Domains}                    & \textbf{ROUGE Scores} & \textbf{BART Fine-tuning} & \textbf{SDPT}  & \multicolumn{1}{l}{\begin{tabular}[c]{@{}l@{}}\textbf{SDPT}\\     \textbf{w/ RecAdam}\end{tabular}} & \textbf{DAPT}  & \multicolumn{1}{l}{\begin{tabular}[c]{@{}l@{}}\textbf{DAPT}\\     \textbf{w/ RecAdam}\end{tabular}} & \textbf{TAPT}  & \multicolumn{1}{l}{\begin{tabular}[c]{@{}l@{}}\textbf{TAPT}\\     \textbf{w/ RecAdam}\end{tabular}} \\ \hline \hline
\multirow{3}{*}{\textbf{\textbf{Dialog}}}    & ROUGE-1 F1   & 39.95            & 42.84 & \textbf{45.23}                                                                             & 41.22 & 40.05                                                                             & 40.15 & 41.34                                                                             \\
                           & ROUGE-2 F1   & 17.50            & 17.51 & \textbf{19.43}                                                                             & 17.88 & 17.62                                                                             & 16.99 & 17.88                                                                             \\
                           & ROUGE-L F1   & 31.64            & 33.79 & \textbf{35.37}                                                                             & 32.40  & 31.36                                                                             & 31.21 & 32.31                                                                             \\ \hline
\multirow{3}{*}{\textbf{Email}}     & ROUGE-1 F1   & 24.71            & 25.16 & \textbf{26.97}                                                                             & 26.50  & 25.66                                                                            & 25.30  & 25.73                                                                             \\
                           & ROUGE-2 F1   & 11.71            & 12.2  & \textbf{13.44}                                                                             & 13.14 & 12.89                                                                             & 12.03 & 12.69                                                                             \\
                           & ROUGE-L F1   & 24.15            & 24.28 & \textbf{25.98}                                                                             & 25.61 & 25.14                                                                             & 24.63 & 25.32                                                                             \\ \hline
\multirow{3}{*}{\textbf{Movie R.}}  & ROUGE-1 F1   & 25.13            & 25.45 & \textbf{26.06}                                                                             & 24.25 & 25.78                                                                             & 25.27 & 25.65                                                                             \\
                           & ROUGE-2 F1   & 9.22             & 9.49  & \textbf{10.27}                                                                             & 9.06  & 9.84                                                                              & 9.24  & 9.13                                                                              \\
                           & ROUGE-L F1   & 20.04            & 20.11 & \textbf{20.91}                                                                             & 19.56 & 20.69                                                                             & 20.09 & 20.45                                                                             \\ \hline
\multirow{3}{*}{\textbf{Debate}}    & ROUGE-1 F1   & 24.48            & 25.61 & 25.17                                                                             & \textbf{26.71} & 25.01                                                                             & 24.59 & 24.70                                                                              \\
                           & ROUGE-2 F1   & 8.21             & 8.48  & 8.38                                                                              & \textbf{9.14}  & 8.42                                                                              & 8.13  & 8.43                                                                              \\
                           & ROUGE-L F1   & 21.96            & 22.86 & 22.39                                                                             & \textbf{23.64} & 22.17                                                                             & 22.04 & 22.25                                                                             \\ \hline
\multirow{3}{*}{\textbf{Social M.}} & ROUGE-1 F1   & 21.76            & 22.43 & \textbf{23.25}                                                                             & 22.95 & 21.51                                                                             & 22.81 & 23.01                                                                             \\
                           & ROUGE-2 F1   & 8.11             & 9.06  & 9.01                                                                              & \textbf{9.66}  & 8.25                                                                              & 8.96  & 8.49                                                                              \\
                           & ROUGE-L F1   & 21.03            & 21.03 & \textbf{22.18}                                                                             & 21.93 & 20.69                                                                             & 22.06 & 21.95                                                                             \\ \hline
\multirow{3}{*}{\textbf{Science}}   & ROUGE-1 F1   & 72.76            & \textbf{73.09} & 72.60                                                                             & 71.88 & 72.23                                                                             & 73.08 & 72.80                                                                              \\
                           & ROUGE-2 F1   & 64.66            & \textbf{65.15} & 63.79                                                                             & 63.73 & 63.32                                                                             & 65.04 & 64.26                                                                             \\
                           & ROUGE-L F1   & 68.40            & 68.62 & 68.06                                                                             & 67.34 & 67.62                                                                             & \textbf{68.81} & 68.41                                                                             \\ \hline
\end{tabular}
\end{adjustbox}
\caption{Full results of all models.}
\label{tab:all_results}
\end{table*}

\section{Training Details}
\label{sec:appendix B}
Our model contains $\sim$139.4 million parameters and we train all models on one GTX 1080 Ti. We train all the models for 50 epochs in around three hours. We manually tune the hyperparameter values. 

\end{document}